\documentclass{article}
\pdfoutput=1
\usepackage{ijcai22}

\usepackage{times}

\usepackage{soul}
\usepackage{url}
\usepackage[utf8]{inputenc}
\usepackage[small]{caption}
\usepackage{graphicx}
\usepackage{amsmath}
\usepackage{booktabs}
\usepackage{multirow}
\usepackage{multicol}
\usepackage[frozencache,cachedir=.]{minted}
\title{WrapperFL: A Model Agnostic Plug-in for Industrial Federated Learning}

\author{
Xueyang Wu$^1$\footnote{Contact Author}\and
Shengqi Tan$^2$\and
Qian Xu$^{1,2}$\And
Qiang Yang$^{1,2}$\\
\affiliations
$^1$The Hong Kong University of Science and Technology, Hong Kong SAR, China\\
$^2$WeBank Inc., Shenzhen, China\\
\emails
xwuba@connect.ust.hk, elontan@webank.com,
qianxu@ust.hk, qyang@cse.ust.hk
}

\begin{document}

\maketitle

\begin{abstract}
Federated learning, as a privacy-preserving collaborative machine learning paradigm, has been gaining more and more attention in the industry. With the huge rise in demand, there have been many federated learning platforms that allow federated participants to set up and build a federated model from scratch. However, existing platforms are highly intrusive, complicated, and hard to integrate with built machine learning models. For many real-world businesses that already have mature serving models, existing federated learning platforms have high entry barriers and development costs. This paper presents a simple yet practical federated learning plug-in inspired by ensemble learning, dubbed WrapperFL, allowing participants to build/join a federated system with existing models at minimal costs. The WrapperFL works in a plug-and-play way by simply attaching to the input and output interfaces of an existing model, without the need of re-development, significantly reducing the overhead of manpower and resources. We verify our proposed method on diverse tasks under heterogeneous data distributions and heterogeneous models. The experimental results demonstrate that WrapperFL can be successfully applied to a wide range of applications under practical settings and improves the local model with federated learning at a low cost.

\end{abstract}

\section{Introduction}
Artificial intelligence (AI) techniques require considerable training data to achieve satisfying performance, especially for industrial applications. Therefore, one of the major challenges in building AI models is collecting enough training data. Conventionally,  the model developers could aggregate data by purchasing from multiple data owners or directly collecting data from the service users. However, these ways raise the risk of data privacy leakage and are forbidden by regulations. For example, the General Data Protection Regulation (GDPR)~\cite{voss2016european} enacted by the European Union (EU) rectifies the usage of personal data and prohibits the transfer of personal data collected from users. The data privacy protection regulations pose a new challenge for AI developers to utilize the scattered data to build AI models. 

To tackle this challenge, researchers propose a new distributed machine learning paradigm, dubbed Federated Learning (FL), which allows multiple participants to jointly train a machine learning model without directly sharing their private data \cite{yang2019federated,mcmahan2017communication}. With a rapid development, FL are gaining widespread applications, such as language modeling \cite{mcmahan2017learning8}, topic modeling \cite{jiang2019federated103,jiang2021industrial}, speech recognition \cite{Jiang2021GDPR}, healthcare~\cite{chen2020fedhealth}, etc. Apart from the development of FL in research areas, the high commercial value of FL also motivates companies and institutions to explore industrial-level FL platforms, such as Tensorflow Federated~\cite{tff}, PaddleFL~\cite{ma2019paddlepaddle}, and FATE~\cite{liu2021fate}, which provide comprehensive infrastructures for businesses to set up their FL prototypes and services. However, the application of such platforms requires the AI developers to be knowledgeable and familiar with not only the federated learning algorithms but also the applied platforms \cite{zhuang2022easyfl}. Therefore, it turns out to be a high barrier for many potential industrial businesses. On the other hand, industrial users usually have their existing AI models that have been sufficiently evaluated on real-world businesses. In contrast, the existing FL algorithms and platforms require the participants to train federated models from scratch, bringing extra overheads and risks to the running AI services. 

Hence, these come up with two consequent requirements for applying FL to practical industrial businesses. First, the insertion of FL should be lightweight, without increasing significant efforts and expert knowledge. Second, the federalization should be pluggable, which means the user can seamlessly switch its models between the federated and the non-federated versions.

\begin{figure*}[!htp]
    \centering
	\includegraphics[width=0.8\linewidth]{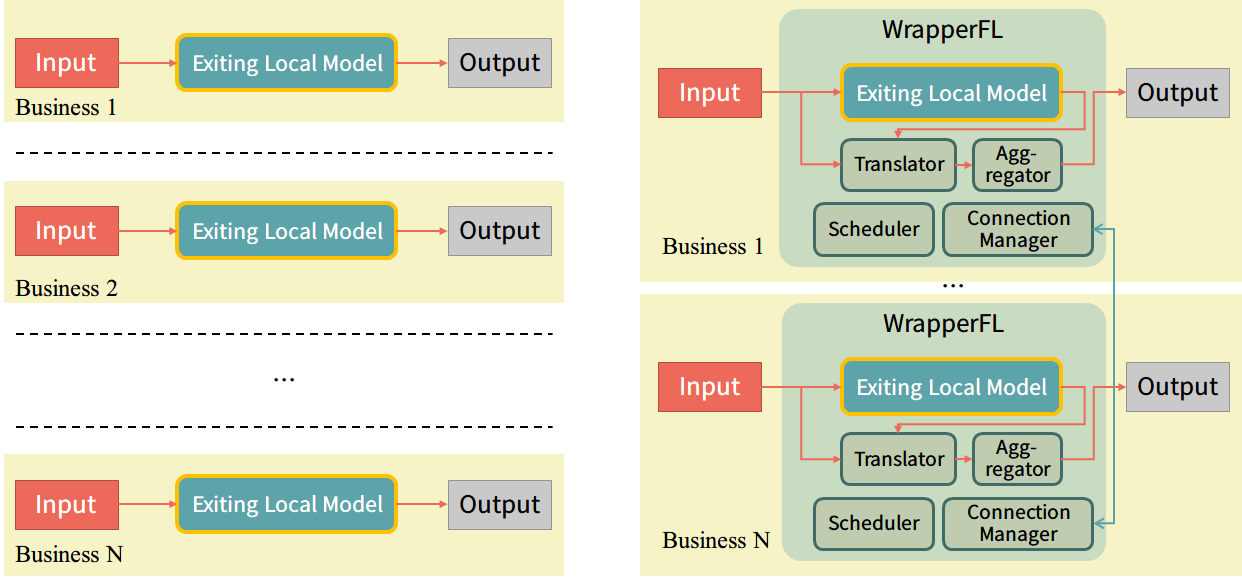}
	\caption{The general framework of WrapperFL. The left part shows the existing local models that are deployed separately in different businesses. The right part depicts the federated system where the existing local models of different businesses are extended with WrapperFL.}
	\label{fig:framework}
\end{figure*}

Motivated by these, this paper proposes a novel, simple yet effective federated learning framework, dubbed WrapperFL, which can be easily plugged into an existing machine learning system with little extra effort. The framework is non-intrusive, simply attaching to and reusing the input and the output interfaces of an existing machine learning model. With the spirit of ensemble learning \cite{sagi2018ensemble}, WrapperFL regards the existing machine learning models as the experts whose knowledge can be exchanged with the others in a federated manner. Different from existing federated learning algorithms, such as FedAvg\cite{mcmahan2017communication},  which train a joint model from scratch, WrapperFL aims to exploit existing models, which further poses another challenge of dealing with the heterogeneity in data distributions and existing models \cite{li2020federated-fedprox,karimireddy2020scaffold}. More specifically, we design two model-agnostic strategies of WrapperFL, i.e., StackingWrapper and BaggingWrapper. The former is suitable for centralized federated learning, where exists a central curator that coordinates all federated participants; the latter is suitable for decentralized federated learning without a central curator. The details are described in section~\ref{sec:method}.

As the name states, the WrapperFL converts an existing non-federated model to the federated version by wrapping it with a highly integrated toolkit, which contains learning-related components, i.e., translator and aggregator, and communication-related components, i.e., scheduler and connection manager. As shown in Figure \ref{fig:framework}, the WrapperFL wraps the existing model by attaching it with external components without changing the original data flow as well as the existing local model. This maintains the usability of the existing model in its production environment. Besides, WrapperFL preserves identical input and output interfaces as the original local model for the upstream and downstream services, avoiding extra costs of adjusting existing services. The scheduler and connection manager are responsible for coordinating the federated system as well as transferring model parameters, which are decoupled from the data flow and interface logic, and are agnostic to existing services.  

To demonstrate the efficacy and feasibility of our proposed WrapperFL, we conduct comprehensive experiments on diverse tasks with different settings, including heterogeneous data distributions and heterogeneous models. The experimental results demonstrate that WrapperFL can be successfully applied to a wide range of applications under practical settings and improves the local model with federated learning. Meanwhile, the cost of switching a non-federated local model to a federated version is low in terms of both developing efforts and training time. 

In summary, the major contribution of this paper is that we propose a pluggable federated learning toolkit, WrapperFL, which can painlessly convert an existing local model to a federated version and achieves performance gain.

\section{Related Work}
\subsection{Federated Learning Platforms}
With the huge demand from academia and industry, developers have proposed several FL platforms. The FL platforms can be roughly categorized into research-oriented and industry-oriented. Tensorflow Federated (TFF)~\cite{tff} and FederatedScope~\cite{federatedscope}, for example, focus on the support of deep learning models and allow flexible FL mechanism designs, which are favorable for FL research. However, they lack flow management and support for vertical federated learning \cite{yang2019federated}, which makes them not feasible for real-world industrial applications. Federated AI Technology Enabler (FATE)~\cite{liu2021fate} is the first open-source industrial FL platform that provides comprehensive APIs and flow controls for constructing federated applications, leading the trend of FL applications and inspiring other platforms such as PaddleFL~\cite{ma2019paddlepaddle}. However, they are not user-friendly due to their complicated system architecture and high barriers. Recently, Zhuang et al. propose EasyFL~\cite{zhuang2022easyfl}, a low-code platform that simplifies the development of FL models and allows for rapid testing. 

Compared to FL platforms, WrapperFL focuses on providing an easy-to-use FL plugin for industrial users with existing models for services. WrapperFL is not a platform to implement or develop new FL algorithms. Therefore, if one wants to propose or invent new FL algorithms, WrapperFL might be inferior. Instead, WrapperFL is a plug-and-play toolkit that allows users to rapidly switch an existing model to the federated version and achieve performance gain. 

\subsection{Ensemble Learning}
The spirit of ensemble learning is to achieve higher prediction performance by combining multiple trained models, and the ensemble result is better than any of the individual model~\cite{sagi2018ensemble}. Ensemble learning provides a feasible solution to utilize existing models from other clients. First, the existing local models are similar to the weak models in ensemble learning, and thus the appropriate fusion of their results could lead to a better prediction. Second, typically, when there is a lot of variation among the models, ensembles get superior outcomes \cite{sollich1995learning,kuncheva2003measures}, implying that heterogeneous local models might bring extra benefits. 

Based on the idea of ensemble learning, WrapperFL adopts two ensemble strategies, stacking \cite{breiman2001random} and bagging \cite{buhlmann2012bagging}. The stacking strategy introduces a shared model across clients that takes the raw data and the output of the local model as the input. The shared model performs like a virtual bridge that connects local models to ensemble their knowledge. The bagging strategy adopts a simple linear regressor to re-weight the outputs from all local models. Notably, WrapperFL is different from conventionally federated learning algorithms that jointly learn a model from scratch; in contrast, WrapperFL aims to utilize the knowledge in other existing models to enhance the prediction of the existing local model.

\section{Methodology}
\label{sec:method}
In this section, we propose the methodology of WrapperFL. Unlike FL platforms, such as Tensorflow Federated~\cite{tff}, FATE~\cite{liu2021fate}, PaddleFL~\cite{ma2019paddlepaddle}, EasyFL~\cite{zhuang2022easyfl}, etc., WrapperFL is not designed to provide a comprehensive and fundamental toolkit for developing federated learning algorithms. In contrast, WrapperFL aims to help businesses achieve performance gains by federating their existing AI models at a minimum cost. Therefore, the WrapperFL does not involve complicated machine learning operators (MLOps), nor does it provides customized flow control. As this paper focuses on the learning strategy used in WrapperFL, we leave the engineering details, such as life cycle and communication, in the extended version. 

\subsection{Framework}
As shown in Figure~\ref{fig:framework}, WrapperFL is attached to the local model by extending to the input and output interfaces. The existing system can regard the WrapperFL (i.e., the light green block in Figure~\ref{fig:framework}) as the original model as WrapperFL does not change the data flow logic. Inside the WrapperFL, the local model is empowered by two specific ensemble learning-based federated strategies, Stacking Wrapper and Bagging Wrapper. In abstraction, WrapperFL contains four components:
\begin{itemize}
    \item \textbf{Translator}: The translator is in charge of aligning heterogeneous models and producing the federated output.
    \item \textbf{Aggregator}: The aggregator fuses the output of the original model and the federated model\footnote{Notably, in this paper, we do not use the term \textit{global model}. Generally, the global model refers to the horizontal federated learning with homogeneous architecture. While in our settings, the federated participants might have distinct local models. Therefore, we use the term federated model to represent the non-local model(s).}.
    \item \textbf{Scheduler}: The scheduler controls the federated learning flow, including the status synchronization, life cycle management, federated update, etc.
    \item \textbf{Connection Manager}: The connection manager contains the functions of network connection, data transmission, authorization, etc. 
\end{itemize}

\subsection{Stacking Wrapper}
\label{sec:stacking}
\begin{figure}[!htp]
    \centering
	\includegraphics[width=\linewidth]{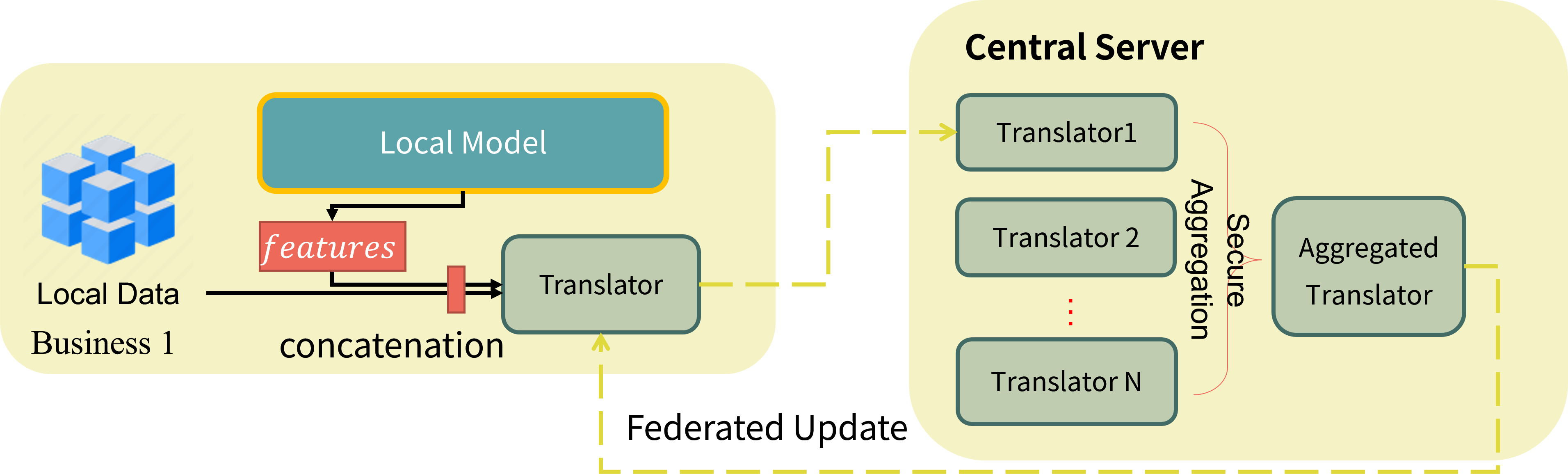}
	\caption{The data flow of Stacking Wrapper.}
	\label{fig:stacking}
\end{figure}

In this section, we introduce details of the stacking strategy for WrapperFL, named Stacking Wrapper. Stacking Wrapper is inspired by the stacking method in ensemble learning that combines the predictions from multiple machine learning models. As Figure \ref{fig:stacking} shows, the Stacking Wrapper essentially stacks two models, the original local model and the translator. The raw data and its feature provided by the existing model (e.g., the hidden vector of the last layer in a neural network) are concatenated as the extended input for the translator. Notably, although the local models could be heterogeneous, the translators are homogeneous across participants, and thus we can apply the Secure Aggregation  \cite{bonawitz2017practical3} to the training of translators. This architecture implicitly conducts the feature space alignment \cite{day2017survey} across different federated participants. 

Stacking Wrapper requires the local models to be parametric models that naturally generate feature vectors. It is not compliant with non-parametric models such as Decision Trees and Support Vector Machines (SVMs).

\subsection{Bagging Wrapper}
\label{sec:bagging}
\begin{figure}[!htp]
    \centering
	\includegraphics[width=0.9\linewidth]{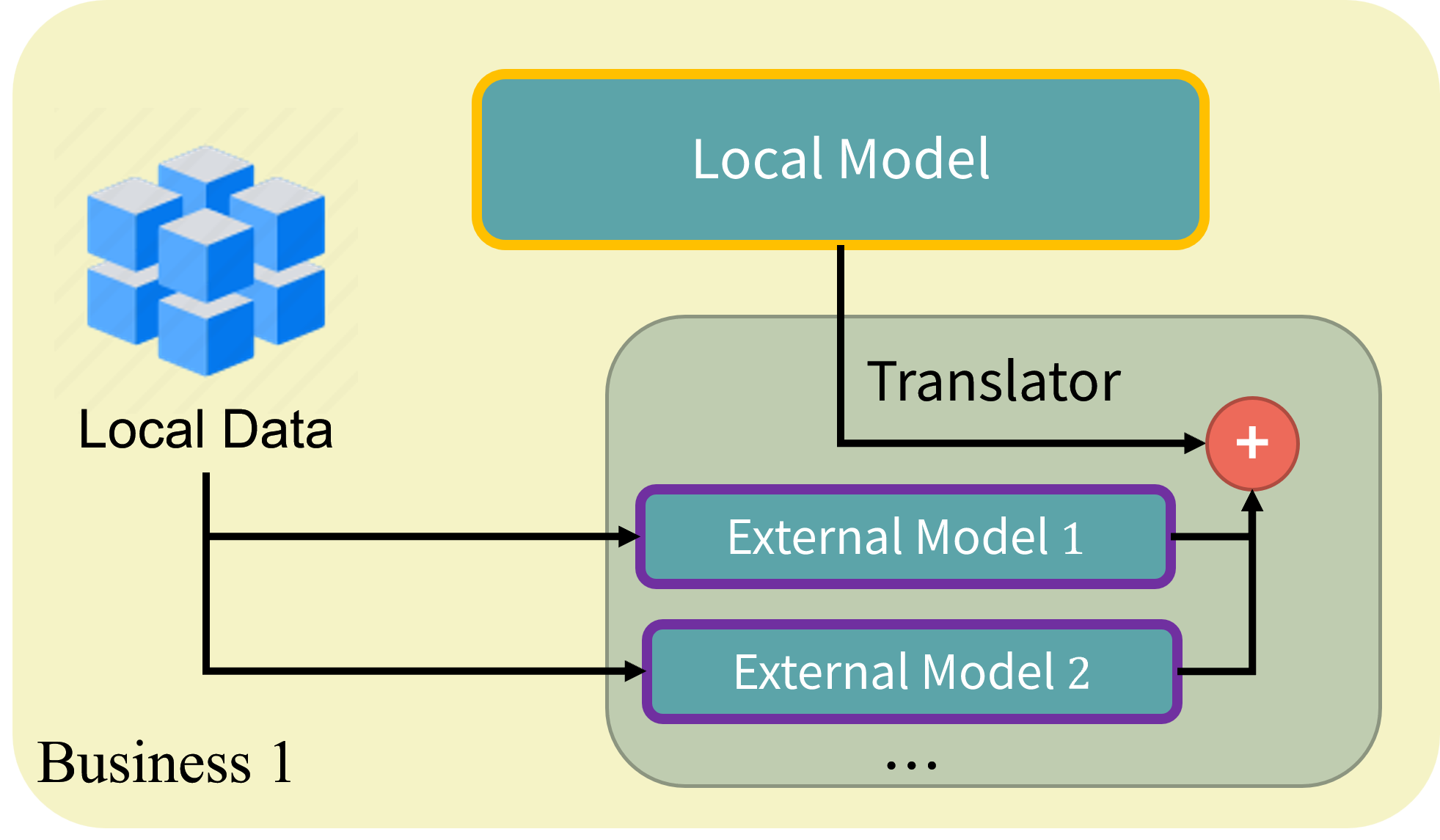}
	\caption{The data flow of Bagging Wrapper.}
	\label{fig:bagging}
\end{figure}

Bagging Wrapper leverages the idea of Bagging in ensemble learning \cite{buhlmann2012bagging}, where we treat the dataset in each client as a random sample from a global data distribution and ensemble the predictions from the local models. As shown in Figure \ref{fig:bagging}, the translator essentially embeds local models received from other participants, infers predictions from all models locally, and fuses the outputs. Compared to Stacking Wrapper, Bagging Wrapper does not pose any assumption on the architecture of local models and thus supports both parametric and non-parametric models. However, a major issue is that transferred local models might raise potential privacy risks. Notably, the risk of privacy leakage from models could be eliminated by incorporating with Trusted Execution Environment (TEE) or Fully Homomorphic Encryption (FHE). 

\begin{table*}[!htp]
\small
\centering
\caption{Results on bank telemarketing prediction of local models and WrapperFL models under imbalanced data settings.}
\label{tab:tab_imb}
\begin{tabular}{rrrrrrrrr}
\toprule
\multicolumn{1}{l}{\multirow{2}{*}{\#Clients}} & \multicolumn{2}{c}{Accuracy}                              & \multicolumn{2}{c}{Precision}                             & \multicolumn{2}{c}{Recall}                                & \multicolumn{2}{c}{F1}                                    \\ \cline{2-9}
\multicolumn{1}{l}{}                           & \multicolumn{1}{c}{Local} & \multicolumn{1}{c}{WrapperFL} & \multicolumn{1}{c}{Local} & \multicolumn{1}{c}{WrapperFL} & \multicolumn{1}{c}{Local} & \multicolumn{1}{c}{WrapperFL} & \multicolumn{1}{c}{Local} & \multicolumn{1}{c}{WrapperFL} \\ \midrule
5  & 0.8888$\pm$0.01 & 0.9032$\pm$0.01 & 0.4489$\pm$0.07 & 0.5998$\pm$0.01 & 0.1530$\pm$0.01 & 0.2846$\pm$0.01 & 0.1980$\pm$0.02 & 0.3831$\pm$0.01\\
10 & 0.8588$\pm$0.01 & 0.9120$\pm$0.01 & 0.1350$\pm$0.06 & 0.6132$\pm$0.05 & 0.1321$\pm$0.04 & 0.3172$\pm$0.02 & 0.1049$\pm$0.02 & 0.4110$\pm$0.03 \\
20 & 0.7991$\pm$0.02 & 0.8983$\pm$0.01 & 0.1048$\pm$0.04 & 0.5145$\pm$0.12 & 0.1864$\pm$0.08 & 0.2556$\pm$0.03 & 0.1096$\pm$0.02 & 0.3215$\pm$0.05\\\bottomrule
\end{tabular}
\end{table*}

\begin{table*}[]
    \small
    \centering
    \caption{Results on bank telemarking prediction of local models and WrapperFL models under non-IID data settings.}
    \label{tab:tab_non}
    \begin{tabular}{rrrrrrrrr}
    \toprule
    \multicolumn{1}{l}{\multirow{2}{*}{\#Clients}} & \multicolumn{2}{c}{Accuracy}                              & \multicolumn{2}{c}{Precision}                             & \multicolumn{2}{c}{Recall}                                & \multicolumn{2}{c}{F1}                                    \\ \cline{2-9} 
    \multicolumn{1}{l}{}                           & \multicolumn{1}{c}{Local} & \multicolumn{1}{c}{WrapperFL} & \multicolumn{1}{c}{Local} & \multicolumn{1}{c}{WrapperFL} & \multicolumn{1}{c}{Local} & \multicolumn{1}{c}{WrapperFL} & \multicolumn{1}{c}{Local} & \multicolumn{1}{c}{WrapperFL} \\ \midrule
    5  & 0.8968$\pm$0.01 & 0.9064$\pm$0.01 & 0.4753$\pm$0.06 & 0.7173$\pm$0.02 & 0.1523$\pm$0.01 & 0.2532$\pm$0.01 & 0.2097$\pm$0.02 & 0.3642$\pm$0.01\\
    10 & 0.8637$\pm$0.01 & 0.9040$\pm$0.01 & 0.2079$\pm$0.05 & 0.6573$\pm$0.03 & 0.1768$\pm$0.03 & 0.3105$\pm$0.02 & 0.1752$\pm$0.03 & 0.3862$\pm$0.02\\
    20 & 0.7967$\pm$0.02 & 0.9056$\pm$0.01 & 0.1286$\pm$0.05 & 0.6592$\pm$0.04 & 0.1769$\pm$0.05 & 0.3504$\pm$0.01 & 0.1052$\pm$0.01 & 0.4375$\pm$0.01\\\bottomrule
    \end{tabular}
\end{table*}

\subsection{Examples of Using WrapperFL}
The WrapperFL is easy to use. To set up a federated client, the user only needs to configure very few parameters, e.g., the existing local model, the client id, client's and server's IP addresses and ports for connection, the type of translator, etc. The WrapperFL will automatically complete the registration, initialization, and building connections with clients. Once instantiating the WrapperFL class with a simple configuration, the user can begin training the federated model with one line of code. Hence, the WrapperFL model can be used the same way as the local model. 

The following code block shows four examples of using WrapperFL, which remarkably demonstrate the ease of use of WrapperFL. Lines 6-12 and 25-30 exemplify Stacking Wrapper and Bagging Wrapper configurations. The user can register to a federated network by filling their identifier in \texttt{client\_id}, local address in \texttt{client\_addr}, server address in \texttt{server\_addr}, and other federated clients' ids in \texttt{clients}. The fields of \texttt{local\_model} and \texttt{train\_dataset} are related to the local model and local training dataset. These are used to train the translator specified in the field \texttt{translator}. 

Apart from the configuration, the user can federalize an existing local model with only three lines of code (i.e., lines 14, 16, and 21), which provides superior convenience for developers compared to other FL toolkits. Notably, in the inference phase, the WrapperFL model performs identically to the original local model (shown in Example 3 and Example 4), significantly reducing the efforts to adjust the existing interfaces. 

{\scriptsize
\begin{minted}[xleftmargin=20pt,linenos]{python}
from wrapperfl import BaggingWrapper, StackingWrapper

## Initialization

# --- Example 1: Federalization with Stacking Wrapper  ---
configs = {"local model": local_model, 
           "train_dataset":dataset, 
           "translator":"LR", # translator structure
           'client_id':'0', 
           'clients':['1', '2', ...], 
           'client_addr':{'ip':'x.x.x.x', 'port':'x'}, 
           'server_addr':{'ip':'x.x.x.x', 'port':'x'}}
# registration and initialization
wrapper = StackingWrapper(configs) 
#train the translator
wrapper.start(status='train', 
              train_config={'local_epochs':10, 
                            'rounds':10, 
                            ...}) 
#swtich to the inference mode
wrapper.start(status='infer', 
              infer_config={'threshold':0.5, ...}) 

# --- Example 2: Federalization with Bagging Wrapper
configs = {"local model": local_model, 
           "train_dataset":dataset,
           'clients':['1', '2', ...], 
           'client_id':'0', 
           'client_addr':{'ip':'x.x.x.x', 'port':'x'}, 
           'server_addr':{'ip':'x.x.x.x','port':'x'}}
wrapper = BaggingWrapper(configs) 
wrapper.start(status = 'train', 
              train_config={'local_epochs':10, ...})) 
wrapper.start(status = 'infer', 
              infer_config={'threshold':0.5, ...}) 

## Usage of WrapperFL model 

# --- Example 3: Sklearn-style local model
# local model inference
y = local_model.predict(X) 
# WrapperFL model inference
y = wrapper.model.predict(X) 

# --- Example 4: PyTroch-style local model
# local model inference
y = local_model(X) 
# WrapperFL model inference
y = wrapper.model(X) 
\end{minted}
}

\section{Experiment}
To evaluate the effectiveness of our proposed WrapperFL, we conduct comprehensive experiments on diverse tasks with different settings, including heterogeneous data distributions and heterogeneous models. 

\subsection{Datasets}
We verify our proposed WrapperFL with three different forms of data, including the bank telemarketing prediction \cite{moro2014bankdata} (tabular data), news classification~\cite{tianchidata} (textual data), image classification with CIFAR-10~\cite{cifar10} (image data). These datasets are large-scale and thus allow us to manipulate their data distributions to form heterogeneous data distributions including \textbf{imbalanced} and \textbf{non-IID} data distributions. Notably, although non-IID data contains imbalanced data in some literature~\cite{zhu2021federated-noniid-survey}, in this paper, we refer \textbf{imbalanced} data to the setting that class distributions across clients are (almost) identical, while \textbf{non-IID} data to the setting that class distributions are different.

To simulate a different number of clients, we partition the tabular data and text data into 5, 10, and 20 clients, and we split the image data into 5 and 10 clients; as in the case of 20 clients, the heterogeneity in data distribution across clients are not apparent. 

To synthesize imbalanced datasets, we first assign the number of instances for each client following the Dirichlet Process~\cite{Teh2010}, and uniformly sample the instances for each client from the global data distribution. For non-IID datasets, we directly assign the number of samples of each class in each client following the Dirichlet Process. A special case is the bank telemarketing prediction data, which has already been class-imbalanced. For this case, we only sample the number of positive samples and then pad the local datasets with negative samples to the identical size. Figure~\ref{fig:tab_dist}-\ref{fig:image_dist} illustrate the overview of class-aware distributions of each client of three datasets, which are shown in appendix \ref{apdx:figure}

\subsection{Evaluation Metrics}
We compute the accuracy, precision, recall, and F1-score for all tasks for each client's local and WrapperFL models under a balanced global testing set. We report the means and standard deviations of the metrics under each setting. 

\subsection{Experiments on Tabular Dataset}
Tabular datasets are common in real-world industries. We evaluate the effectiveness of WrapperFL on the bank telemarketing prediction data~\cite{moro2014bankdata} with heterogeneous model structures. 

More specifically, we set 40\% of the clients to use Logistic Regression (LR) as the local modes, and 60\% of the clients to use the three-layer MLP with various hidden sizes (16, 18, and 24 each for 20\% of the clients). Table~\ref{tab:tab_imb} shows the performance comparison of the local models and our WrapperFL models under imbalanced data settings. In this case, the WrapperFL adopts a three-layer MLP with the hidden size of 16 as the translator, which is trained with ten communication rounds using FedAvg~\cite{mcmahan2017communication}. 

Notably, when the number of clients increases, the precision of local models drops dramatically due to the highly imbalanced local data. In contrast, extended with WrapperFL, the participants achieve significantly better performance. For example, when there are ten clients, the mean F1 score of WrapperFL models is 2.92x better than that of local models. We can make similar conclusions in the case of non-IID data settings, which are shown in Table~\ref{tab:tab_non}. In this case, the WrapperFL model shows strong robustness against the local models. For example, the mean F1-score of WrapperFL is 3.16x higher than that of local models in the setting of 20 clients.

\begin{table*}[]
    \small
    \centering
    \caption{Results on news classification of local models and WrapperFL models under imbalanced data settings.}
    \label{tab:text_imb}
    \begin{tabular}{rrrrrrrrr}
    \toprule
    \multicolumn{1}{l}{\multirow{2}{*}{\#Clients}} & \multicolumn{2}{c}{Accuracy}                              & \multicolumn{2}{c}{Precision}                             & \multicolumn{2}{c}{Recall}                                & \multicolumn{2}{c}{F1}                                    \\ \cline{2-9} 
    \multicolumn{1}{l}{}                           & \multicolumn{1}{c}{Local} & \multicolumn{1}{c}{WrapperFL} & \multicolumn{1}{c}{Local} & \multicolumn{1}{c}{WrapperFL} & \multicolumn{1}{c}{Local} & \multicolumn{1}{c}{WrapperFL} & \multicolumn{1}{c}{Local} & \multicolumn{1}{c}{WrapperFL} \\ \midrule
    5  & 0.2129$\pm$0.08 & 0.2836$\pm$0.05 & 0.1647$\pm$0.10 & 0.1743$\pm$0.08 & 0.1968$\pm$0.06 & 0.2085$\pm$0.05 & 0.1507$\pm$0.07 & 0.1655$\pm$0.06\\
    10 & 0.2003$\pm$0.08 & 0.2311$\pm$0.07 & 0.1438$\pm$0.07 & 0.1305$\pm$0.06 & 0.1726$\pm$0.04 & 0.1753$\pm$0.04 & 0.1273$\pm$0.05 & 0.1264$\pm$0.05\\
    20 & 0.2140$\pm$0.03 & 0.3977$\pm$0.01 & 0.0720$\pm$0.03 & 0.4032$\pm$0.01 & 0.1160$\pm$0.01 & 0.2911$\pm$0.01 & 0.0703$\pm$0.02 & 0.2514$\pm$0.01
    \\\bottomrule
    \end{tabular}
\end{table*}

\begin{table*}[]
    \small
    \centering
    \caption{Results on news classification of local models and WrapperFL models under non-IID data settings.}
    \label{tab:text_non}
    \begin{tabular}{rrrrrrrrr}
    \toprule
    \multicolumn{1}{l}{\multirow{2}{*}{\#Clients}} & \multicolumn{2}{c}{Accuracy}                              & \multicolumn{2}{c}{Precision}                             & \multicolumn{2}{c}{Recall}                                & \multicolumn{2}{c}{F1}                                    \\ \cline{2-9} 
    \multicolumn{1}{l}{}                           & \multicolumn{1}{c}{Local} & \multicolumn{1}{c}{WrapperFL} & \multicolumn{1}{c}{Local} & \multicolumn{1}{c}{WrapperFL} & \multicolumn{1}{c}{Local} & \multicolumn{1}{c}{WrapperFL} & \multicolumn{1}{c}{Local} & \multicolumn{1}{c}{WrapperFL} \\ \midrule
    5  & 0.6259$\pm$0.01 & 0.6977$\pm$0.01 & 0.4355$\pm$0.01 & 0.4892$\pm$0.01 & 0.3833$\pm$0.01 & 0.4035$\pm$0.01 & 0.3580$\pm$0.01 & 0.3962$\pm$0.01\\
    10 & 0.5533$\pm$0.01 & 0.6561$\pm$0.01 & 0.3056$\pm$0.01 & 0.3643$\pm$0.01 & 0.3119$\pm$0.01 & 0.3404$\pm$0.01 & 0.2815$\pm$0.01 & 0.3221$\pm$0.01\\
    20 & 0.5152$\pm$0.01 & 0.6288$\pm$0.01 & 0.2157$\pm$0.01 & 0.3457$\pm$0.01 & 0.2597$\pm$0.01 & 0.3069$\pm$0.01 & 0.2128$\pm$0.01 & 0.2912$\pm$0.01
    \\\bottomrule
    \end{tabular}
\end{table*}

\subsection{Experiments on Text Dataset}

Text data is also commonly used in the industry~\cite{devlin2018bert55}. We select the news classification dataset~\cite{tianchidata} to represent the NLP tasks as text classification is one of the most important tasks in NLP. Unlike the tabular data, the news classification dataset has 14 evenly distributed classes. Moreover, the feature distribution in news data (i.e., the word distributions) is more consistent across different classes. Therefore, it requires the classifier to capture the deep semantics of a sentence. 

\begin{table}[!htp]
\small
    \caption{Results on image classification of local models and WrapperFL models under imbalanced data settings.}
    \label{tab:img_imb}
    \begin{tabular}{ccrr}
        \toprule
        \multicolumn{2}{l}{\#Clients}          & \multicolumn{1}{c}{5} & \multicolumn{1}{c}{10} \\ \midrule
        \multirow{2}{*}{Accuracy}  & Local     & 0.7444$\pm$0.0049 & 0.6301$\pm$0.0182  \\ \cline{2-2}
                                   & WrapperFL & 0.7798$\pm$0.0018 & 0.8254$\pm$0.0004  \\\midrule
        \multirow{2}{*}{Precision} & Local     & 0.7412$\pm$0.0055 & 0.6340$\pm$0.0153  \\ \cline{2-2}
                                   & WrapperFL & 0.7823$\pm$0.0016 & 0.8284$\pm$0.004   \\ \midrule
        \multirow{2}{*}{Recall}    & Local     & 0.7443$\pm$0.0049 & 0.6304$\pm$0.0177  \\ \cline{2-2}
                                   & WrapperFL & 0.7793$\pm$0.0018 & 0.8258$\pm$0.0004  \\ \midrule
        \multirow{2}{*}{F1}        & Local     & 0.7416$\pm$0.0054 & 0.6233$\pm$ 0.0218 \\ \cline{2-2}
                                   & WrapperFL & 0.7781$\pm$0.0019 & 0.8241$\pm$0.0005\\ \bottomrule
        \end{tabular}
\end{table}

In this experiment, we evaluate the compatibility of WrapperFL with different deep learning models. We randomly assign 20\% of the clients with TextCNN~\cite{jacovi2018understanding}, 40\% of the clients with Transformer \cite{vaswani2017attention} (with three stacking encoders), and 40\% of the clients with Transformer (with two stacking encoders) as their local models. WrapperFL adopts a single-encoder Transformer as the translator, which is trained with ten communication rounds using FedAvg~\cite{mcmahan2017communication}.

Table \ref{tab:text_imb} and Table \ref{tab:text_non} display the performance comparisons between local models and WrapperFL models, from which we can confirm the high utility of WrapperFL on complex deep learning models. For example, in the setting of 20 clients, the WrapperFL models perform much better than local models no matter of accuracy, precision, recall, or F1 score. 

\subsection{Experiments on Image Dataset}
\label{sec:exp:img}
Processing image data is also a huge requirement in the industry, such as face recognition. We adopt one of the most widely used image datasets, CIFAR-10~\cite{cifar10}, to verify our WrapperFL on CNN models with different scales. We randomly assign 20\% of clients with ResNet18~\cite{he2016deep}, 40\% of clients with VGG13~\cite{simonyan2014very}, and 40\% of clients with VGG16. WrapperFL adopts a much smaller CNN model, VGG11, as the translator, which is trained with ten communication rounds using FedAvg~\cite{mcmahan2017communication}. 

Table~\ref{tab:img_imb} and Table~\ref{tab:img_non} display the performances of local models and WrapperFL models. Consistently, we can observe significant performance gains comparing the WrapperFL models to the local models. Moreover, the experimental results also demonstrate that using a much smaller homogeneous translator can efficiently and effectively distill knowledge from well-trained large-scale local models.

\begin{table}[]
\small
    \caption{Results on image classification of local models and WrapperFL models under non-IID data settings.}
    \label{tab:img_non}
    \begin{tabular}{ccrr}
        \toprule
        \multicolumn{2}{l}{\#Clients}          & \multicolumn{1}{c}{5} & \multicolumn{1}{c}{10} \\ \midrule
        \multirow{2}{*}{Accuracy}  & Local     & 0.6902$\pm$0.0020 & 0.5747$\pm$0.0031 \\ \cline{2-2} 
                                   & WrapperFL & 0.7405$\pm$0.0026 & 0.7853$\pm$0.0015 \\ \hline
        \multirow{2}{*}{Precision} & Local     & 0.6617$\pm$0.0066 & 0.5563$\pm$0.0055 \\ \cline{2-2} 
                                   & WrapperFL & 0.7511$\pm$0.0016 & 0.8017$\pm$0.0005 \\ \hline
        \multirow{2}{*}{Recall}    & Local     & 0.6877$\pm$0.0020 & 0.5737$\pm$0.0027 \\ \cline{2-2} 
                                   & WrapperFL & 0.7404$\pm$0.0025 & 0.7862$\pm$0.0012 \\ \hline
        \multirow{2}{*}{F1}        & Local     & 0.6628$\pm$0.0042 & 0.5377$\pm$0.0048 \\ \cline{2-2} 
                                   & WrapperFL & 0.7249$\pm$0.0053 & 0.7792$\pm$0.0020 \\ \bottomrule
        \end{tabular}
\end{table}

\subsection{Quantitative Comparison of Training Cost}
While it is hard to quantify the reduction in the required manpower and resources of WrapperFL compared to existing FL platforms, we can empirically validate its feasibility. Most importantly, WrapperFL utilizes the existing local model, significantly reducing the training time and tuning the local model, which is well-known for both time and labor-consuming. Besides, with a simple plug-and-play package, ML developers without background knowledge of federated learning can quickly adapt their services to the federated version, which eliminates the barrier of manpower of FL experts.  

\begin{figure}[!htp]
    \centering
	\includegraphics[width=0.9\linewidth]{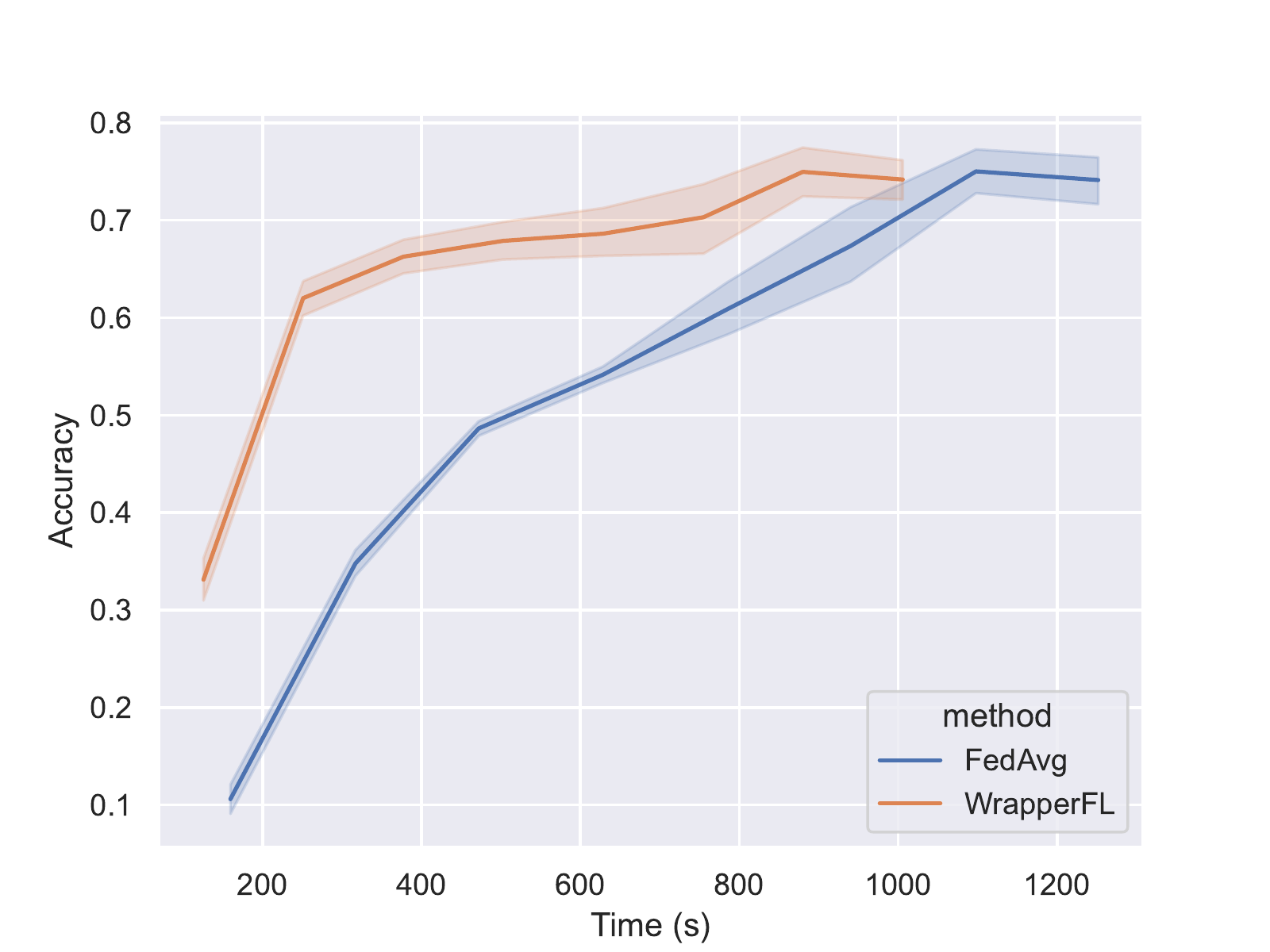}
	\caption{The Accuracy Curves of WrapperFL and FedAvg on Image Dataset}
	\label{fig:time_cost}
\end{figure}

To further show the training cost (including the time cost and the potential performance loss), we conduct a quantitative comparison between WrapperFL and the FedAvg on the non-IID dataset mentioned in Section \ref{sec:exp:img}. In Figure~\ref{fig:time_cost}, we plot the accuracy points at training time intervals of WrapperFL and FedAvg, respectively, from which we can observe that WrapperFL achieves over 60\% accuracy much faster than FedAvg. Moreover, we can also observe that the performance of WrapperFL is comparable to FedAvg when converged, which indicates that performance loss of WrapperFL is negligible. Notably, the dataset used in this experiment is much smaller than industrial-scale data and the training time reported here ignores the time cost of deployment and communication. Therefore, in practice, the time acceleration of WrapperFL could be much more significant. 

\section{Conclusion}
This paper focus on the problem of applying federated learning to the industry, where the businesses have already had a mature and existing model and have no experts on federated learning algorithms and platforms. The problem is challenging from two aspects. First, the federated learning process should be completely decoupled from the existing system, and the federalization should be lightweight, low-intrusive, and require as less development workload as possible. Second, the federated learning toolkit should be model-agnostic. Therefore, it can be compatible with various model architectures, and the users can seamlessly change the underlying machine learning models. To tackle these challenges, we propose a simple yet practical federated learning plug-in inspired by ensemble learning, dubbed WrapperFL, allowing participants to build/join a federated system with existing models at minimal costs. The experimental results of tabular, text, and image data demonstrate the efficacy of our proposed WrapperFL on both model heterogeneous and data heterogeneous settings.

\onecolumn
\newpage

\appendix
\section{The Details of Data Distributions}
\label{apdx:figure}

\begin{figure*}[!htp]
    \centering
	\includegraphics[width=0.7\linewidth]{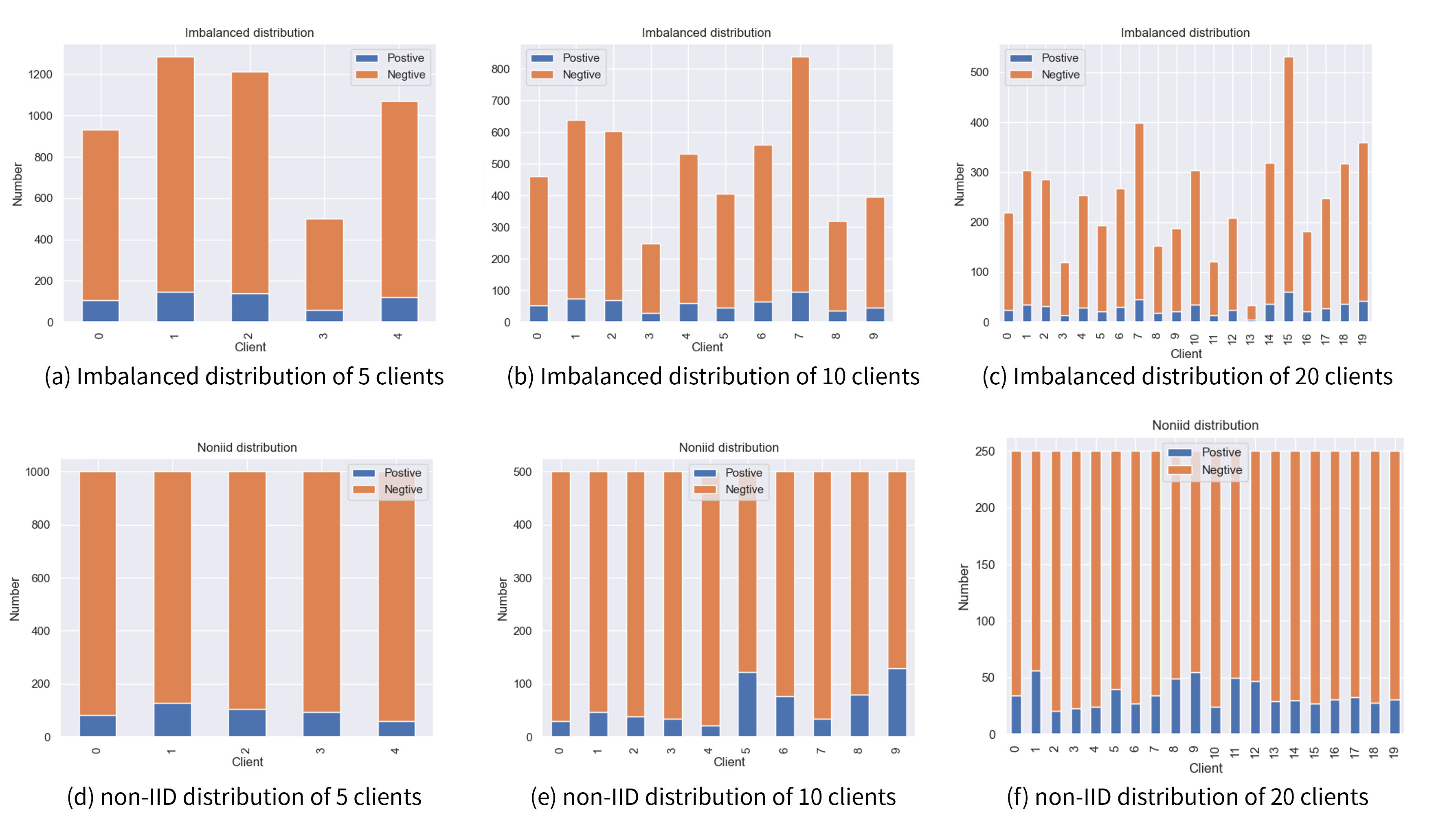}
	\caption{The stacking histograms of distributions of the bank telemarketing prediction data.}
	\label{fig:tab_dist}
\end{figure*}

\begin{figure*}[!htp]
    \centering
	\includegraphics[width=0.7\linewidth]{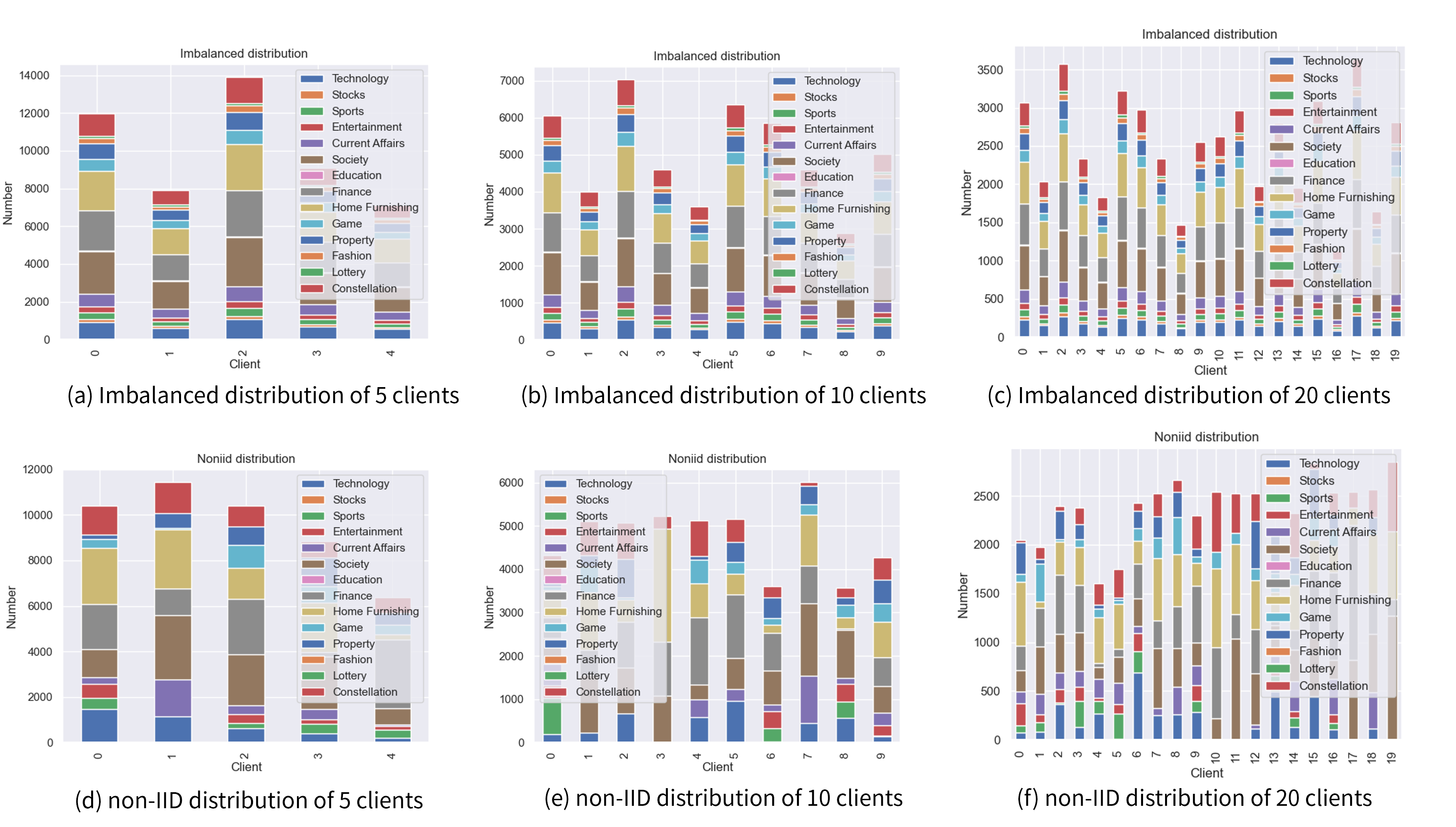}
	\caption{The stacking histograms of distributions of the news classification data.}
	\label{fig:text_dist}
\end{figure*}

\begin{figure*}[!htp]
    \centering
	\includegraphics[width=0.8\linewidth]{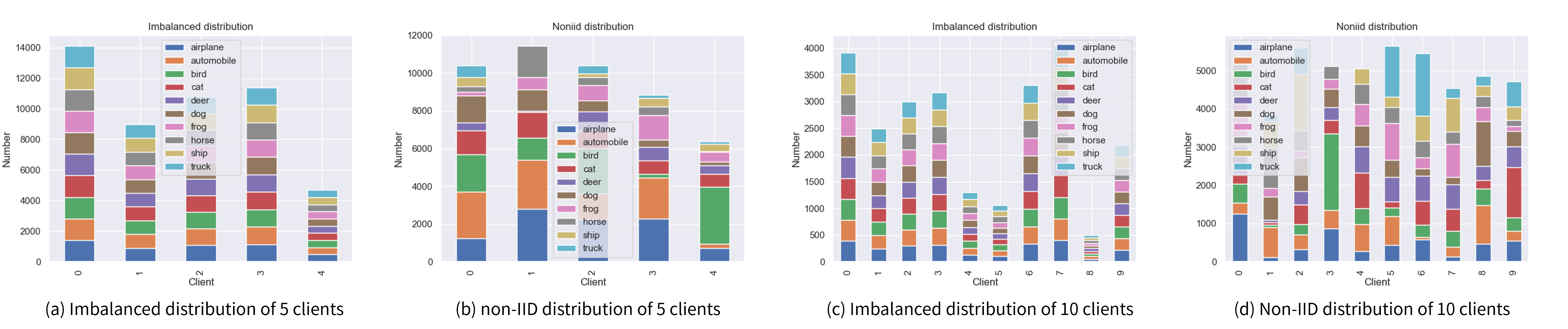}
	\caption{The stacking histograms of distributions of the image classification data.}
	\label{fig:image_dist}
\end{figure*}

\end{document}